\documentclass[letterpaper, 10 pt, conference]{ieeeconf}

\IEEEoverridecommandlockouts                              
\usepackage{amsmath,amsfonts}
\usepackage{amssymb}
\usepackage{array}
\usepackage{graphicx}
\usepackage[utf8]{inputenc}
\usepackage{xurl}
\usepackage{caption}
\usepackage[hidelinks]{hyperref}

\DeclareMathOperator*{\argmax}{arg\,max}

\usepackage{booktabs}
\usepackage{orcidlink}

\title{\LARGE \bf Reinforcement Learning for Ballbot Navigation in Uneven Terrain} 
\author{Achkan Salehi $^{\dagger}$ \thanks{\noindent $\dagger$ VimaLabs, France. Email: \texttt{achkan@vimalabs.com}. All work in this paper was completed independently prior to joining VimaLabs.}\orcidlink{0000-0002-7180-0123}} 

\begin{document}
\maketitle
\thispagestyle{empty}
\pagestyle{empty}

\begin{small}

\begin{abstract}
  Ballbot (Ball balancing robot) navigation relies on methods rooted in control theory (CT), and works that apply Reinforcement Learning (RL) to the problem remain rare while being limited to specific subtasks. Unlike CT based methods, RL does not require simplifying assumptions about environment dynamics. In addition to this increased accuracy in modeling, RL agents can easily be conditioned on observations such as depth-maps without the need for explicit formulations from first principles, leading to increased adaptivity. Despite those advantages, there has been little to no investigation into the capabilities, data-efficiency and limitations of RL for ballbot control and navigation. Furthermore, there is a notable absence of an open-source, RL-friendly simulator for this task. We present an open-source ballbot simulation based on MuJoCo, and show that with appropriate conditioning and reward shaping, classical model-free RL methods can produce policies that effectively navigate randomly generated uneven terrain, using a reasonable amount of data (four to five hours on a system operating at 500hz). To our knowledge, none of the current CT based methods are able to tackle such scenarios. Our code is available at \color{blue}\url{https://github.com/salehiac/OpenBallBot-RL}\color{black}.
\end{abstract}

\section{Introduction}
\label{sec_intro}

  The term Ballbot (\textit{i.e.} Ball balancing robot) refers to a family of dynamically stable, underactuated robots that are similar in principle to an inverted pendulum mounted on a spherical base \cite{nagarajan2014ballbot,lauwers2006dynamically,fankhauser2010modeling}. The single spherical base results in omnidirectional mobility and agility/maneuverability, especially in comparison with more conventional, statically stable wheeled robots. However, ballbot control is notoriously challenging due to several factors such as the system's high non-linearity and sensitivity to perturbations \cite{fankhauser2010modeling,fischer2024closed,xiao2023design,pham2024analysis}. Almost all approaches to Ballbot control in the literature are based on methods from control theory that require a mathematical model of the ballbot. Those models generally make several simplifying assumptions \cite{zhou2021learning, karachalios2024parameter, pham2024analysis}, \textit{e.g.} that the ball moves in a perfectly horizontal plane, that there is no slippage between the floor and the ball and that there is no deformation.

  In contrast, Reinforcement Learning (RL) algorithms in general do not require a mathematical model of the robot, and therefore do not require any of the aforementioned simplifications. Furthermore, in an RL context, policies can easily be conditioned on additional observations such inputs from depth cameras, which can result in more adaptive navigation. Despite those advantages, there has been little to no investigation into the capabilities, data-efficiency and limitations of RL based methods for ballbot control and navigation. Furthermore, there is a notable absence of an open-source, RL-friendly simulator for this task.

\begin{figure}[!t]
  \centering
  \includegraphics[width=79mm,clip]{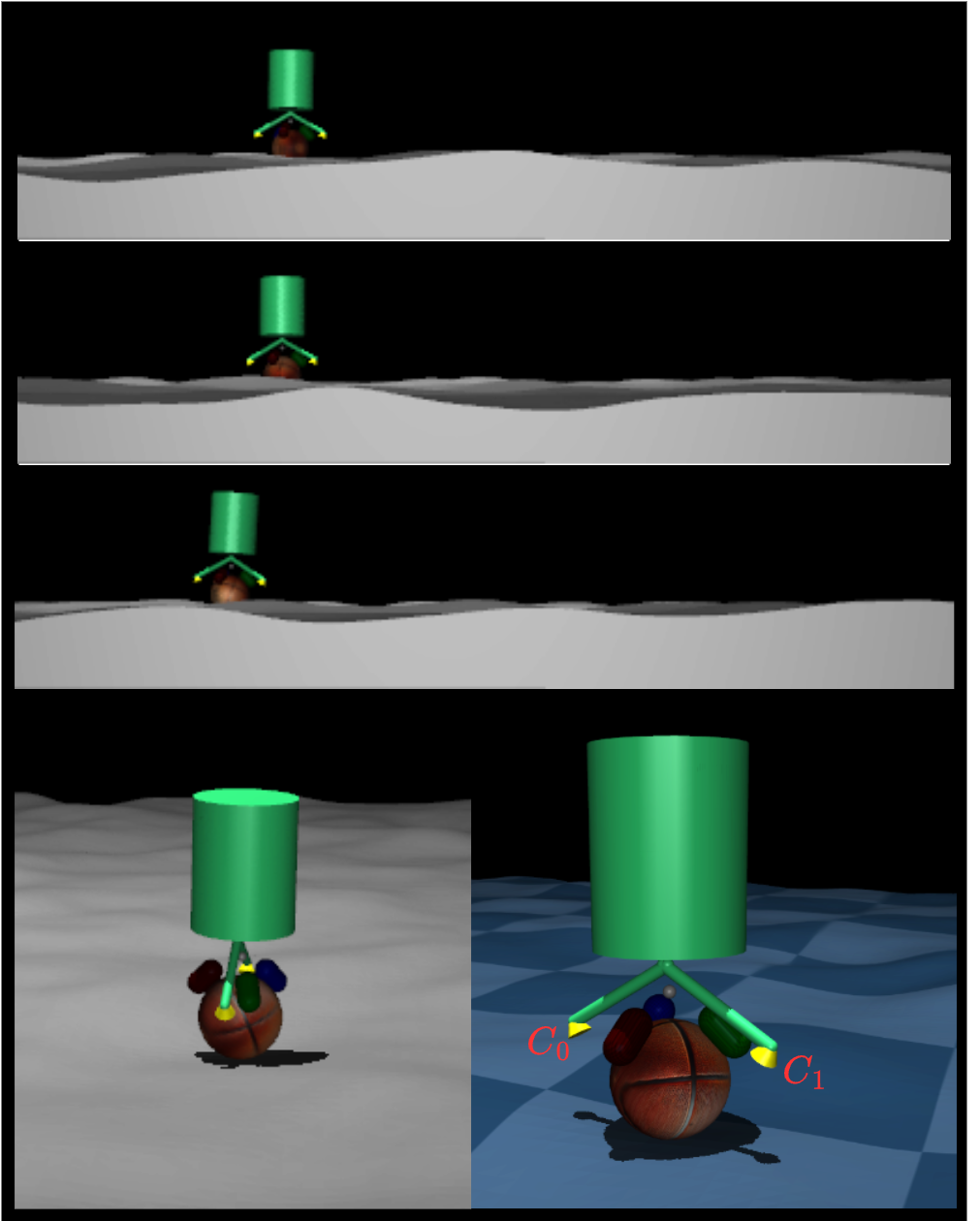}
  \caption{\small{Screenshots from our open-source simulation, where the learned policy navigates through randomly generated uneven terrain. Note that for simplicity, the three omniwheels controlling the base sphere are modeled as capsules with anisotropic tangential friction. Two low-resolution depth cameras (visible as yellow cones and noted $C_0, C_1$ in the bottom right image) enable terrain perception. They are both oriented towards the contact point between the ball and the ground.}}
  \label{fig_first}
\end{figure}

  Our aim in this paper is to demonstrate the potential of Reinforcement Learning for ballbot control/navigation. To this end, we consider the task of navigation in randomly generated uneven terrain, and show that appropriately shaped rewards coupled with conditioning on exteroceptive information allow classical model-free RL methods to succeed in learning an adaptive and robust agent that generalizes to unseen terrain. Figure \ref{fig_first} shows a number of randomly generated terrains as well as our agent navigating those environments. To our knowledge, none of the current control-theory based methods are able to tackle such scenarios.

  The simulation environment, which is made available in the public project repository, is based on the high-fidelity MuJoCo \cite{todorov2012mujoco} physics engine. This choice was motivated by the fact that it offers a good balance between speed and physical fidelity, and that agents trained in MuJoCo have been shown to have high transferability \cite{kaup2024review}. 

  The paper is structured as follows: the next section is dedicated to preliminaries and problem definition. We present our RL-based approach to ballbot control/navigation in \S\ref{sec_method}, and Section \S\ref{sec_sim} discusses our simulated environment. Experiments are presented in section \S\ref{sec_exp}, and are followed by a discussion in \S\ref{sec_discussion}. We review the related literature in \S\ref{sec_related}. Concluding remarks are the subject of section \S\ref{sec_concl}.

\section{Preliminaries and problem formulation}
  \label{sec_background}

  We will first briefly introduce the ballbot and give a high-level overview of the principles underlying most approaches to its control. We then formalize our objectives using Reinforcement Learning terminology.

  \subsection{The ballbot}
  A ballbot is an underactuated, statically unstable but dynamically stable system, similar in principle to an inverted pendulum balancing on a spherical base. While ball drive mechanisms such as inverse mouse-ball drives \cite{nagarajan2014ballbot} and spherical induction motors \cite{seyfarth2016initial} have been used, the most popular drive for ballbots is arguably the omniwheel based setup, an example of which is given in figure \ref{fig_cad}. An omniwheel applies torque to the sphere along its direction of rotation, while its free-rolling idler rollers allow unconstrained slip along the wheel’s axis, enabling omnidirectional motion.

  Ballbot control methods in general define $q\triangleq [x,y, \theta_0, \theta_1, \theta_2]^T$ (where $[x,y]^T$ is the planar position and the $\theta_i$ specify the body's orientation) and use the Euler-Lagrange Equation 

  \begin{equation}
    \frac{d}{dt}\frac{\partial L}{\partial {\dot{q}}} - \frac{\partial L}{\partial q}=\tau_{ext}
  \end{equation}

  from which one can derive the equations of motion as 

  \begin{equation}
    M(q)\ddot{q}+C(q,\dot{q})+g(q)=\tau_{ext}
  \end{equation}

  where $M, C$ are respectively the Mass and Coriolis matrices and $g$ is the gravity vector (additional terms such as friction can sometimes be added \cite{zhou2021learning}). These matrices are then estimated depending on the chosen ballbot model (\textit{e.g.} decoupled motion in $xy$ and $xz$ planes \cite{nagarajan2014ballbot, fischer2024closed} or 3d \cite{nagarajan2013shape} etc) and then balance/stabilizing/trajectory tracking is done via different combinations of PID and LQR controllers (and related methods) at different levels. Our aim in this paper is to make abstraction of those mathematical models and their underlying assumptions such as the absence of slippage or planar-only motion. Therefore, these control methods will not be discussed any further.

  \subsection{Problem formulation}
  Our aim in this work is to enable the ballbot to navigate in uneven terrains, which we sample from a Perlin noise distribution $\mathcal{P}_{perlin}$ \cite{perlin2002improving,gustavson2005simplex}. Some example terrains sampled from this distribution can be seen in figures \ref{fig_first} and \ref{fig_trajs_single}. 

  Let $<\mathcal{S},\mathcal{A},\mathcal{P},\mathcal{P}_0,\mathcal{R},\gamma>$ denote a Markov Decision Process (MDP) with $\mathcal{S},\mathcal{A}$ respectively the set of states and actions, $\mathcal{P}(s'|s,a)$ the transition probabilities, $\mathcal{P}_0$ the initial state distribution, $\mathcal{R}(s,a,s')$ the reward function, and $\gamma\in[0,1)$ the discount factor. Let us also denote $J(\theta)=E_{\tau\sim p_{\theta}}[\sum_t^{\infty} r_t\gamma^t]$ the discounted cumulative rewards with the expectation taken over trajectories $\tau$ sampled from the policy $\pi_{\theta}\triangleq p_{\theta}(a|s)$.

  Our objective can then be formalized as designing the appropriate $\mathcal{S}, \mathcal{R}$ and policy architecture to enable the ballbot to learn to navigate in terrains sampled from $\mathcal{P}_{perlin}$, and subsequently solving $\argmax_{\theta} J(\theta)$ to obtain the policy.

\begin{figure}[h!]
  \centering
  \vspace*{0.6cm}
  \includegraphics[scale=0.29,clip]{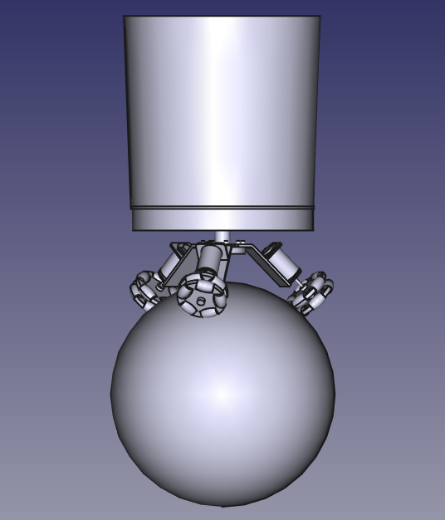}
  \caption{\small{A CAD model of a ballbot that uses three omniwheels as its ball drive mechanism.}}
  \label{fig_cad}
\end{figure}

\begin{figure}[!t]
  \centering
  \includegraphics[width=73mm,trim={0cm 0 0cm 0.0cm},clip]{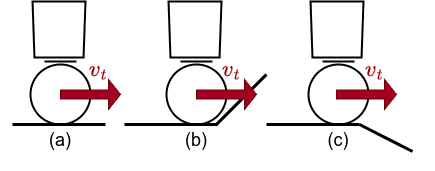}
  \caption{\small{Using only proprioceptive observations in uneven terrain leads to ambiguities: assuming that the robots in (a), (b), (c) have the same state at time $t$ (in the figure, only the velocity $v_t$ is explicitly shown for clarity) and adopting a model-based view, it is clear that any learned model of the form $s_{t+1}=M(a_t,s_t)$ will be ambiguous, as the state information is not sufficient for predicting whether or not the robot will encounter a flat terrain or an increasing/decreasing slope at $t+1$. This ambiguity can not be fully resolved by recurrent models or distributional methods based only on proprioceptive data, but can be alleviated by incorporating observations from exteroceptive sensors, such as depth cameras.}}
  \label{fig_ambig}
\end{figure}

\section{Proposed solution}
 \label{sec_method}

In this section, we will first discuss the choice of state/observation space. This will be followed by the definition of our reward signal, policy architecture and finally training procedure.

Throughout the rest of the paper, the set of proprioceptive observations at timestep $t$ will be noted $o^{prop}_t\triangleq(\phi_t, \dot{\phi_t}, v_t, \dot{m_t}, a_{t-1})$. Here, $\phi_t\in\mathbb{R}^3, \dot{\phi_t}\in\mathbb{R}^3, v_t\in\mathbb{R}^3$ respectively denote orientation, angular velocity and velocity in world coordinates. Each component of the $m_t\in\mathbb{R}^3$ vector denotes the angular velocity of the three omniwheels in local coordinates (\textit{i.e.} relative to their rotation plane). The vector $a_{t-1}\in[0,1]^3$ is the last set of commands sent to each of the three motors.\\

 \noindent\textbf{Observation space.} First, note that naively applying RL to a state defined by $o^{prop}_t$ is sufficient for learning a policy that navigates in a flat plane (the option to train policies in flat terrain, based only on proprioceptive observations, is available in our codebase). However, using only proprioceptive observations leads to ambiguity in uneven terrain (figure \ref{fig_ambig}) and therefore to high uncertainty over actions\footnote{While the use of a terrain distribution might be reminiscent of the task distribution considered in meta-learning \cite{beck2023survey,salehi2023adaptive}, it should be noted that meta-RL methods are applicable when the physical properties and/or reward functions of a novel task can be estimated with sufficient accuracy from earlier data (\textit{e.g.} from a short exploration phase) . In our case, the problem is one of lack of observability that can not be solved by simply using the history of proprioceptive observations.}. In order to alleviate that problem, we endow the robot with the ability to observe the terrain via two low-resolution ($128\times 128$) depth cameras (figure \ref{fig_simu}, left), angled towards the contact point between the robot and the floor. In order to increase efficiency, we do not incorporate those images directly into the state. Instead, we pass them to a pretrained encoder and retrieve the low-dimensional embeddings $z^1\in\mathbb{R}^{20}$ and $z^2\in\mathbb{R}^{20}$.

 A ballbot's control and proprioception usually operate at a high frequency $f_{high}$ that can not be matched by most camera systems. To provide unambiguous observations at the same frequency as $f_{high}$, we use the most recent depth images $z^1_{t'},z^2_{t'}$ at each timestep and append the time difference $\Delta_t=(t-t')$ to the observations. The full observation vector is then defined by

 \begin{equation}
   o_t=(o^{prop}_t, z^1_{t'}, z^2_{t'}, \Delta_t)^T\in\mathbb{R}^{56}
 \end{equation}

 where $t'\leq t$.\\
  
  \noindent\textbf{Policy Network.} The policy architecture used in this paper is illustrated in figure \ref{fig_policy}. Depending on whether the aforementioned depth encoders are trained/fine-tuned or frozen, they can be considered part of the policy. Details can be found in appendix \ref{app_net}.\\
  
  \noindent\textbf{Reward shaping} Noting $g\in \mathbb{R}^2$ the desired direction of motion during an episode, we define our reward signal as

  \begin{equation}
    \mathcal{R}(s_t,a_t)= \alpha_1 v_t^Tg + \mathbf{1}_{\{s_t \notin S_F\}} \alpha_2 - \alpha_3||a_t||^2 
    \label{eq_rew}
  \end{equation}

  where $v_t \subset s_t$ is the velocity at time $t$, the symbol $\mathbf{1}$ denotes the indicator function, $\alpha_i\in\mathbb{R}$ are constants, and $S_F$ is the set of failure states. In this work, a state is considered a failure state if and only if the robot's tilt from the vertical axis is larger than $20^\circ$. Intuitively, the first term in the above rewards the ballbot for moving in the desired direction, the second term gives a positive feedback to the policy for each timestep where it "survives", and the last term acts as regularization by penalizing large motor commands. Hyperparameters used in this paper are reported in appendix \ref{app_train}.\\

\begin{figure}[h!]
  \centering
  \includegraphics[scale=0.52,clip]{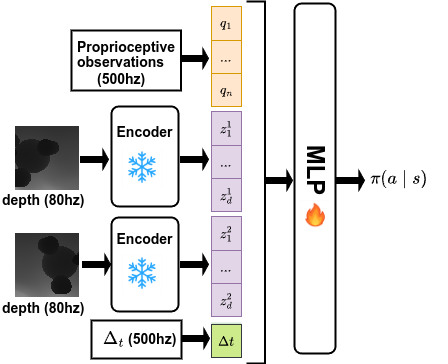}
  \caption{\small{The policy used to navigate uneven terrain. Two low resolution $128\times128$ depth images, one from each depth camera, are fed to a pretrained encoder that maps each of them to an embedding in $\mathbb{R}^{20}$. These embeddings are then concatenated to the proprioceptive observations: orientation, angular velocity, body velocity, angular velocities of each omniwheel (the latter are in local coordinates), as well as the last command vector sent at the previous timestep. To avoid state ambiguities arising from the differences in observation frequencies ($500$hz for proprioceptive readings vs $\sim 80hz$ for the depth cameras), we also concatenate the time $\Delta t$ that has elapsed since the last depth image observation was received. This $56$ dimensional vector is then fed into a small MLP which is trained to predict torque commands that are sent to the omniwheel motors.}}
  \label{fig_policy}
\end{figure}
  
\noindent\textbf{Training.} We used the well-known Proximal Policy Optimization (PPO) algorithm \cite{schulman2017proximal}. While not as data-efficient as other actor-critic methods such as SAC\cite{haarnoja2018soft} or model-based approaches such as DreamerV3 \cite{hafner2023mastering}, it is compute-efficient, has decent exploration capacity (thanks to its entropy term) and its performance is less sensitive to hyperparameter values, making it ideal for an initial simulation-based proof of concept. 

\section{Simulated environment}
 \label{sec_sim}

  The simulation environment (available in the public project repository), is based on the high-fidelity MuJoCo \cite{todorov2012mujoco} physics engine. This choice was motivated by the fact that it offers a good balance between speed and physical fidelity, and that agents trained in MuJoCo have been shown to have high transferability \cite{kaup2024review}. 

  Figure \ref{fig_simu} (left) shows our simulated ballbot, including the two $128\times128$ depth cameras that are oriented towards the contact point between the ball and the ground. As shown in figure \ref{fig_simu} (right), omniwheels are modeled as capsules with anisotropic friction. Note that as of this writing, \textbf{\textit{The current stable MuJoCo release requires a small modification to allow anisotropic friction to function as intended. This tweak is provided as a patch in our public repository, and was suggested by the developers of MuJoCo}}\footnote{See the discussion at \url{https://github.com/google-deepmind/mujoco/discussions/2517}}. 

  Note that robot control is performed at $500hz$, and while proprioceptive observations $o^{prop}_t$ are read at that same frequency, the depth cameras operate at $\sim 80hz$. As discussed in section \S\ref{sec_method}, unambiguous observations $o_t$ are provided at $500hz$ by concatenating the relative timestamp $\Delta t$ to the depth image encodings and proprioceptive readings composing the state.

  The distribution $\mathcal{P}_{perlin}$ is based on the \texttt{snoise2} function provided by \cite{duncan2015noise}\footnote{We set its parameters to \scriptsize{\texttt{scale=25.0},\texttt{octaves=4},\texttt{persistence=0.2},\\ \texttt{lacunarity=2.0} in this work.}}.

\begin{figure}[h!]
  \centering
  \includegraphics[scale=0.37]{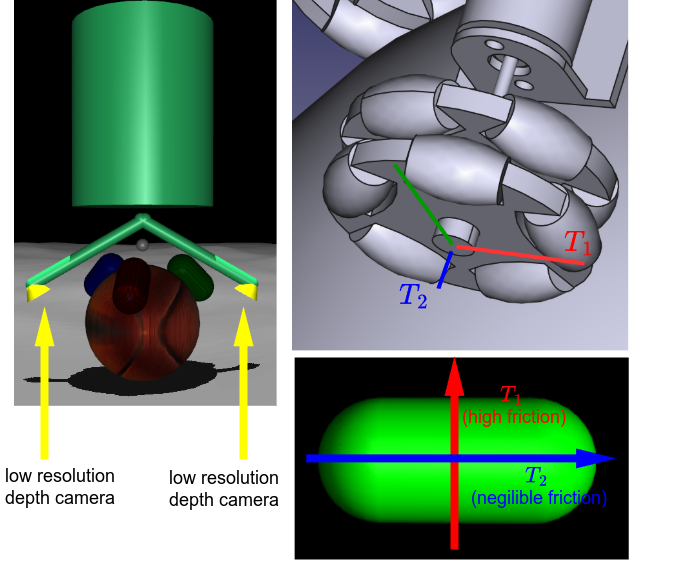}
  \caption{\small{\textbf{(left)} The ballbot from our open-source simulation. \textbf{(Right)} The omniwheels are modeled as capsules with \textit{anisotropic} tangential friction. Friction is high along the $T_1$ axis, which is the direction along which the wheel applies torque to the sphere. Friction in the $T_2$ direction, which corresponds to the rotation direction of the omniwheel's idler rollers, should be negligible.}}
  \label{fig_simu}
\end{figure}

\section{Experiments}
 \label{sec_exp}

 In this section, we:

 \begin{itemize}
   \item show that our trained policies have effectively learned to navigate in uneven terrain and generalize to unseen situations;
   \item provide some qualitative results (more of which can be found in the project's public repository); and
   \item use comparisons with a classical baseline from control theory to assess the difficulty of the uneven terrains from $\mathcal{P}_0$.
 \end{itemize}
 Regarding the last point, we emphasize that ballbot navigation in uneven terrain is an extremely under-explored area, and that we are not aware of any existing method that would serve as a reasonable comparative baseline. As discussed in previous sections (\textit{e.g.} \S\ref{sec_background}), control theory based methods, aside from limiting observations to proprioceptive ones, make the assumption that motion is perfectly planar. As a result, any comparison between those approaches and ours will be unfair. That being said, the performance of algorithms based on control theory can be used to assess the difficulty of the terrain distribution $\mathcal{P}_0$ and verify that they are not too similar to the flat terrain case.\\

 \noindent\textbf{Learning and generalization.} During training, the maximum length of each episode was fixed to a horizon of $H_{train}\triangleq 4000$ steps, with each episode ending either in case of failure (\textit{i.e.} the robot falling down) or after reaching $H_{train}$ steps. We conducted five training experiments, each with a total training budget of $8e6$ environment steps (one step taking $\sim 2ms$ in simulation time) and a different RNG seed.

\begin{figure}[h!]
  \centering
  \includegraphics[scale=0.16,clip]{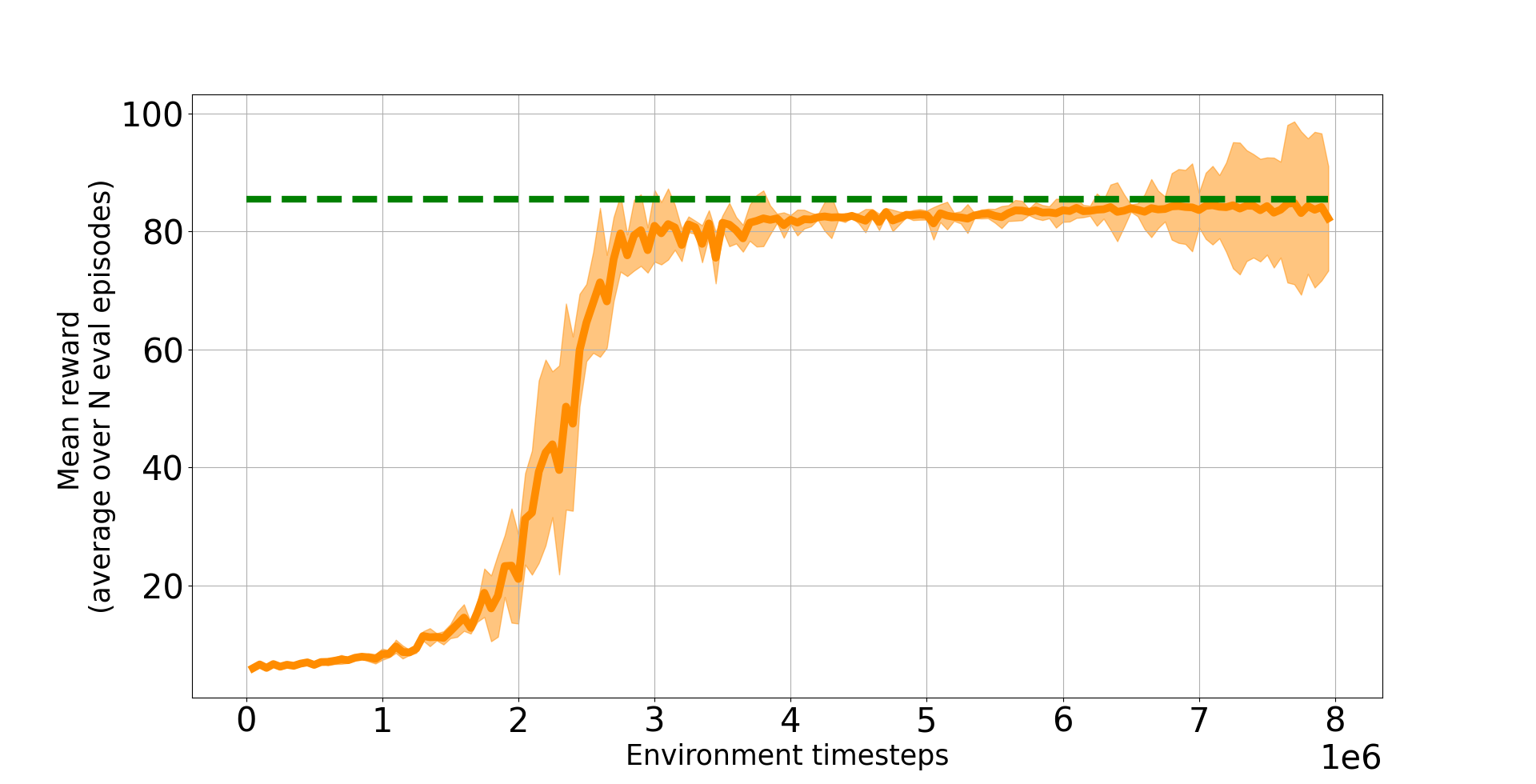}
  \includegraphics[scale=0.16,clip]{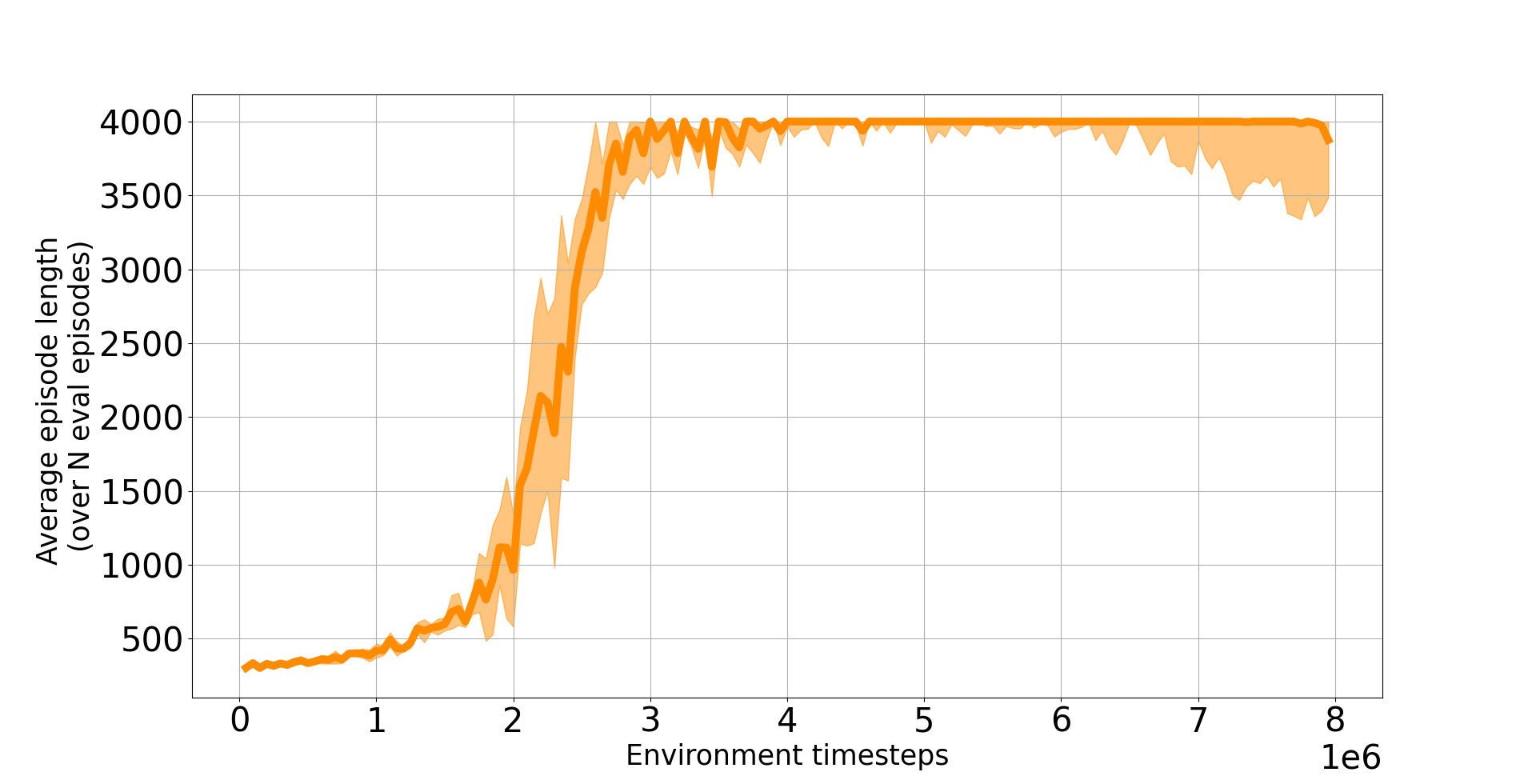}
  \caption{\small{Evolution of the average rewards and episode lengths on \textit{evaluation} (\textit{i.e.} previously unseen) environments during training, from five different training sessions with five different RNG seeds. Each evaluation is based on $N=10$ environments. \textbf{(top)} Median and standard deviation of the average rewards. The green dashed line indicates the \textit{median} of the maximum average reward reached during the five sessions, and is approximately $\sim 85.5$. This reward implies a speed of roughly $0.5m/s$ in the desired direction. \textbf{(b)} The evolution of the average episode lengths. The maximum episode length allowed during training episodes was $4000$.}}
  \label{fig_eval_results}
\end{figure}

\begin{figure}[h!]
  \centering
  \includegraphics[scale=0.2,clip]{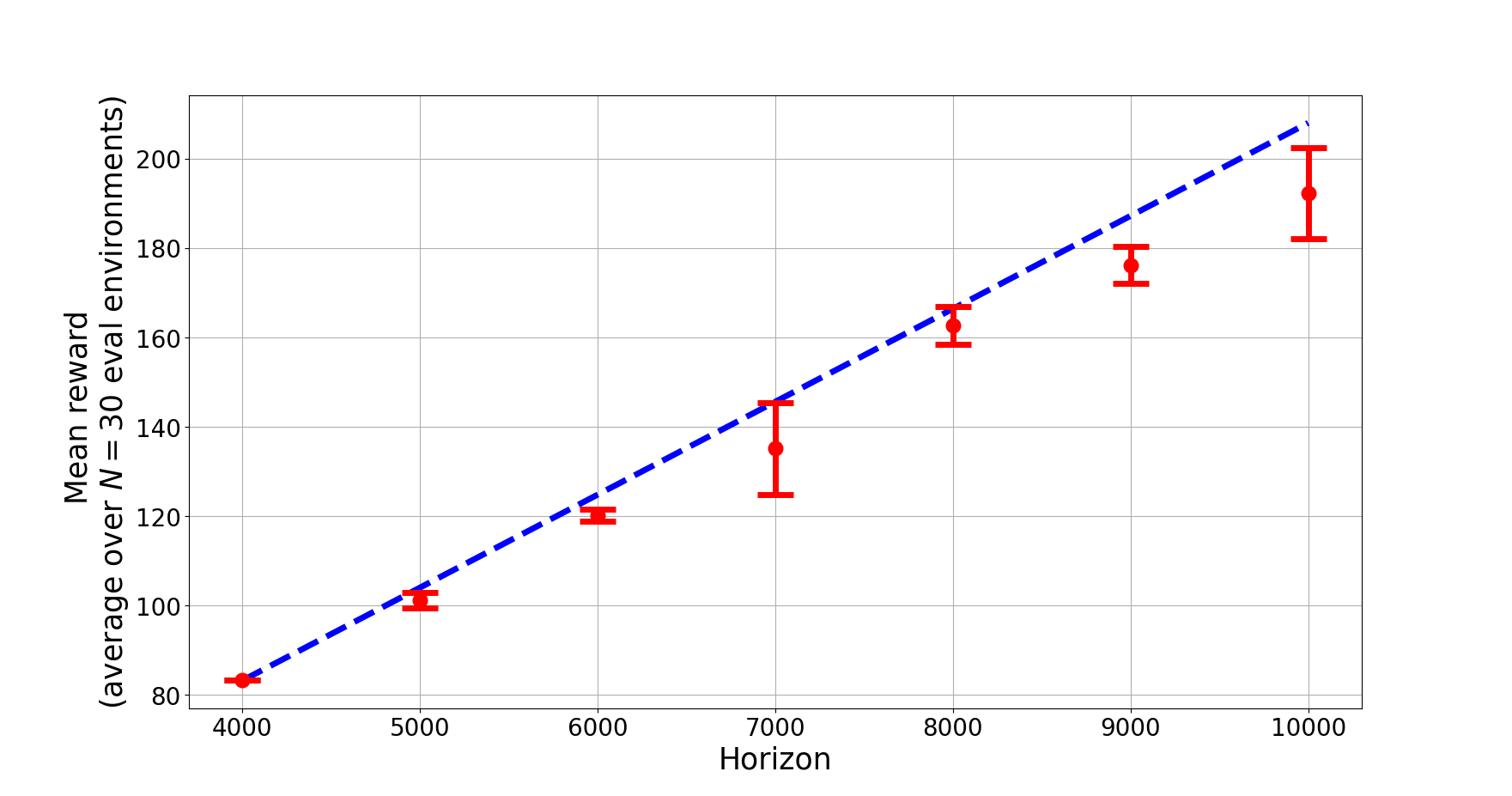}
  \caption{\small{Policies learned within a horizon of $4000$ steps generalize well to longer horizons, as the average reward scales linearly with maximum episode length. Note that each data point $(h,r)$ in the figure results from averaging policy performance over $30$ random environments with horizon $h$, and repeating the experiment with five different RNG seeds.}}
  \label{fig_scale}
\end{figure}

\begin{figure}[h!]
  \centering
  \includegraphics[scale=0.23,clip]{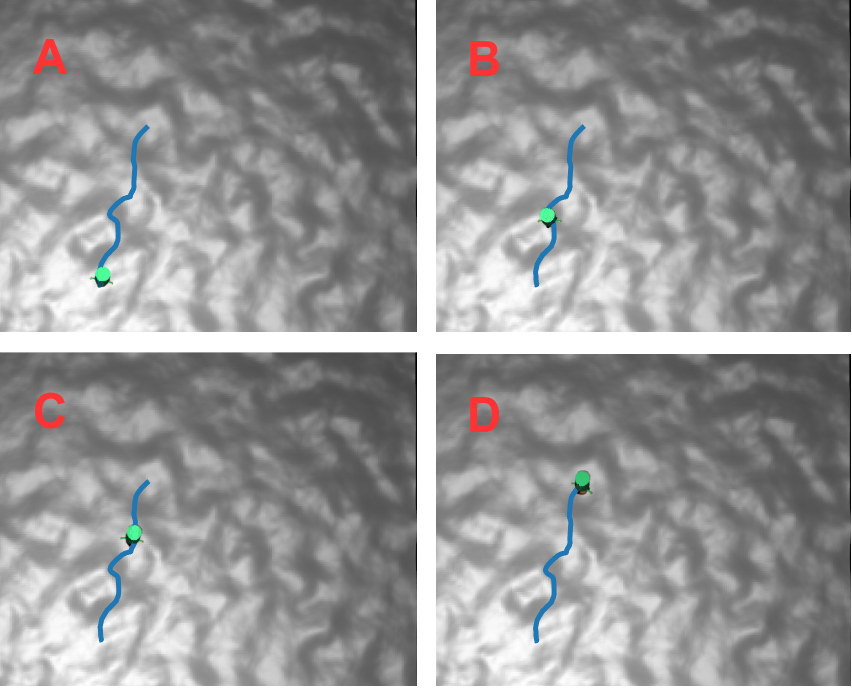}
  \caption{\small{Top down view from a single navigation episode. The trajectory followed by the robot is illustrated in blue. Note that the task is to move in the general $(0,1)$ direction, which coincides with the upwards direction in the image. It can be seen that the robot adapts its navigation to the terrain, circumventing risky areas such as sharp local terrain maxima or minima.}}
  \label{fig_trajs_single}
\end{figure}

\begin{figure}[h!]
  \centering
  \includegraphics[scale=0.23,clip]{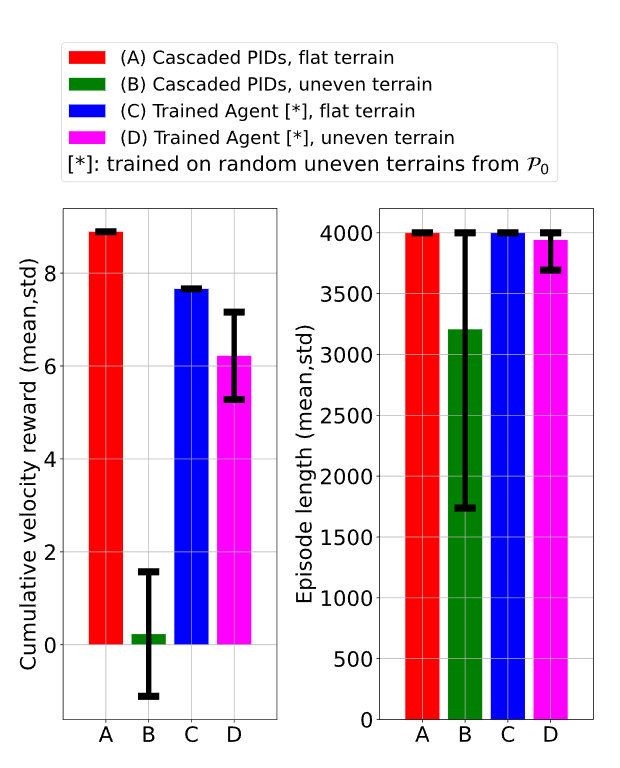}
  \caption{\small{Comparison between cascaded PIDs and a trained policy, on a horizon of $4000$ timesteps.}}
  \label{fig_cascade_comp}
\end{figure}

 The generalization performance of the policy was evaluated at regular intervals, using $N=10$ randomly sampled, previously unseen terrains. The results are reported in figure \ref{fig_eval_results}. In each training experiment, the policy has learned to robustly balance the robot in about $3e6$ timesteps. This is evident from the fact that the average length of evaluation episodes (figure \ref{fig_eval_results}, bottom) starts to plateau at $\sim4000$ environment steps, which is the maximum that is reachable in our settings (shorter episodes would mean the robot has fallen). At this time, the policy has primarily learned to optimize for the second and third terms of equation \ref{eq_rew}. During the remainder of each experiment (environment timesteps $\sim 3e6$ to $8e6$), the remaining reward term (goal direction) is optimized, until the reward plateaus slightly below a value of $86.0$. Noting $r_{max}^{eval}(i)$ the maximum of average evaluation rewards reached during an experiment $i$, $\texttt{median}(\{r_{max}^{eval}(i)\}_{i=1}^{5})\approx 85.5$, which given our choice of hyperparameters for the reward function (appendix \ref{app_train}), translates to an average speed of $\sim 0.5m/s$ in the desired goal direction.

 To further asses generalization, we investigated how the average cumulative reward scales beyond the $4000$ steps horizon over which the policies were trained. More precisely, policies trained within the $4000$ timesteps limit were deployed in previously unseen environments with horizons of $4000+\psi$, with $\psi\in\mathbb{N}$ (we sampled $30$ environments for each value of $\psi$). As shown in figure \ref{fig_scale}, the average rewards gathered by the policies scale almost linearly with horizon, indicating that learned policies do indeed generalize.\\
 
 \noindent\textbf{Qualitative results.} An example of policy behavior in a previously unseen terrain is given in figure \ref{fig_trajs_single} for qualitative illustration. More qualitative examples are available in our repository.\\
 
 \noindent\textbf{Terrain difficulty.} We assessed the difficulty of the terrains sampled from $\mathcal{P}_0$ by observing the performance of cascaded PIDs, which can be considered as a standard baseline for ballbot control \cite{nagarajan2013shape, fankhauser2010modeling}. As is usual, the inner loop of our cascaded PIDs was responsible for balancing, while the outer loop (producing the roll and pitch setpoints for the inner loop) was responsible for reaching the target velocity in the $[0, 1]^T$ direction. The cascaded PIDs were tuned for flat terrain and we compared them to a policy trained on uneven terrain in both flat and uneven environments. The results, which are based on a number of $100$ randomly sampled terrains and a horizon of $4000$, are summarized in figure \ref{fig_cascade_comp}. Note that only velocity rewards (\textit{i.e.} first term of equation \ref{eq_rew}) are taken into account for these comparisons. As shown in figure \ref{fig_cascade_comp} (right), in flat terrain, both policies balance the robot for the maximum possible horizon, while the cascaded PIDs reach higher velocity rewards than our trained policy (figure \ref{fig_cascade_comp} (left)). We hypothesize that this is because the trained policy has \textit{not} been trained on flat terrain, and that it therefore moves in a more conservative way due to the risk of losing balance in uneven terrain. 

 However, the performance of cascaded PIDs plummets on uneven terrain: while the robot is able to keep its balance for $\sim 3000$ timesteps on average \textemdash with considerable standard deviation \textemdash as shown in figure \ref{fig_cascade_comp} (right), it is unable to navigate in the desired direction (figure \ref{fig_cascade_comp} (left)), often navigating in random directions, reaching an almost null reward on average, while only collecting negative rewards in some episodes (as illustrated by the standard deviation in figure \ref{fig_cascade_comp} (left)). This indicates that the terrains sampled from $\mathcal{P}_0$ are not too close to the planar case, and that navigating them is indeed challenging. As shown in both plots from figure \ref{fig_cascade_comp}, our trained policy is only slightly less effective in uneven terrains from $\mathcal{P}_0$ than in flat terrain.

\section{Discussion}
\label{sec_discussion}

Up to this point, we have demonstrated that with appropriate reward shaping and a sufficiently informative observation space, model-free RL algorithms, in particular PPO, can learn policies that allow the ballbot to navigate in uneven terrain and that generalize well to unseen situations. Interestingly, the $8e6$ transitions that were used to obtain those results are equivalent to $\sim4.4$ hours of data, which would be reasonable on a real system. Notice however that we have not aimed for data-efficiency, and that this aspect is likely to be improved by using other actor-critic methods (\textit{e.g.} SAC) or model-based methods such as DreamerV3\cite{hafner2023mastering}. 

While those observations set expectations for what can be expected on a real system and may encourage practitioners to experiment with RL methods, a more promising direction is to transfer policies learned in simulation to real ballbots. While our choice to base the simulation on MuJoCo was partly motivated by its physical high fidelity and high transferability \cite{kaup2024review}, evaluating the sim2real performance of the trained agents is left for future work.

\section{Related work}
\label{sec_related}

The large body of work surrounding ballbot control/navigation, is, as discussed in sections \S\ref{sec_intro} and \S\ref{sec_background}, rooted in control theory. Seminal research on ballbots \cite{lauwers2006dynamically, fankhauser2010modeling, nagarajan2013shape, nagarajan2014ballbot} as well as recent works (\textit{e.g.} \cite{fischer2024closed,zhou2021learning}) are based on mathematical models of the ballbot, which make simplifying assumptions such as perfectly planar motion, the absence of slippage between the floor and the ball, and several other ones regarding friction, elasticity, damping, and so on. In contrast with those methods, we aim for a data-driven approach that is not bound by these assumptions. We note in passing that while Zhou \textit{et al.} leverage RL, they only target balance recovery. Furthermore, the use of RL in their work is limited to improving a conventional feedback controller, which still requires a mathematical model of the ballbot.

Several ballbot simulation environments have been developed by different authors (\textit{e.g.} \cite{jo2020contact,zhou2021learning,song2023hands,nashat2023ballbot}, however, to our knowledge no functional open-source ballbot simulations have been published\footnote{A few repositories exists on github, but seem to have been abandoned mid-development.}. Furthermore, simulations developed as part of prior research projects often rely on physics engines that, unlike MuJoCo, are not ideal for RL research, mainly due to their computational costs and the difficulty of running parallel environments. 

Our research is also related to works in Reinforcement Learning and control that consider operation in difficult terrain \cite{kumar2021rma,zakka2025mujoco, carius2022constrained}. However, algorithms that require adapting the policy based on a short history of observations and actions \cite{kumar2021rma} are not suitable for the ballbot, due to its sensitivity to erroneous controls (the same can be said of meta-learning like approach that require adaptation at deployment). Furthermore, works in this category generally target humanoids or more common embodiments \cite{zakka2025mujoco}. An exception is the work by Carius \textit{et al.}, which considers ballbot control on uneven terrain as a benchmark to evaluate their proposed approach. However, their observations are only proprioceptive, and their policy operates without any observations about the terrain. Naturally, this leads to lower generalization and additional failure modes compared to our approach.

\section{Conclusion}
\label{sec_concl}

Prior work has largely overlooked the possibility of applying RL algorithms to ballbot control, despite the fact that their data-driven nature allows them to operate without a mathematical model of the robot, and therefore to avoid unrealistic or limiting assumptions. Furthermore, to our knowledge, no open-source ballbot simulation is available. We have addressed those issues in this work by 1) providing an open-source, RL-friendly simulated ballbot environment based on MuJoCo and 2) demonstrating that model-free RL algorithms, in particular PPO, can train agents that allow the ballbot to navigate in randomly sampled uneven terrain. To our knowledge, none of the previous ballbot control methods have shown this ability. Future directions include improving data-efficiency and evaluating sim2real transferability.

%
%
%
%

\bibliographystyle{IEEEtran}
\bibliography{example}  
\appendices
\section{Network architecture}
\label{app_net}

The encoder is a vanilla CNN whose details are given in table \ref{table_CNN}. The $56$ dimensional feature vector resulting from concatenating proprioceptive observations and depth embeddings (figure \ref{fig_policy}) is then passed into a small MLP with $5$ linear layers with hidden dimension $128$, LeakyReLu and no normalization.

\begin{table}[!ht]
  \centering
  \caption{\small{Encoder architecture}}
  \label{table_CNN}
  \begin{tabular}{@{}lccc@{}}
    \toprule
    \textbf{Layer} & \textbf{params} \\
    \midrule
    Conv1         & in=1, out=32, kernel sz=3, stride=2, padding=1\\
    BatchNorm     & default \\
    Leaky ReLu    & default \\
    Conv2         & in=32, out=32, kernel sz=3, stride=2, padding=1\\
    BatchNorm     & default \\
    Leaky ReLu    & 0.01 \\
    Flatten       & N.A \\
    Linear & num\_output\_features=20 \\
    BatchNorm & default \\
    Tanh & N.A \\
    \bottomrule
  \end{tabular}
\end{table}

\section{Training details}
\label{app_train}

The coefficient from the reward function (equation \ref{eq_rew}) were set to $\alpha_1=0.01, \alpha_2=0.02, \alpha_3=0.0001$. We used the PPO implementation from \texttt{stable\_baselines3} \cite{stable-baselines3}. The hyperparameters used for training are given in table \ref{table_ppo}. As is usual with PPO, we found that tuning the clip range, entropy coefficient and epochs per updates was critical. Starting with a LR of $1e-4$, we manually scheduled the LR by dividing it by constant factors after a specific number of environment timesteps were reached.

\begin{table}[!ht]
  \centering
  \caption{\small{PPO Hyperparameters}}
  \label{table_ppo}
  \begin{tabular}{@{}lc@{}}
    \toprule
    \textbf{Hyperparameter} & \textbf{Value} \\
    \midrule
    Discount factor ($\gamma$) & 0.99 \\
    GAE parameter ($\lambda$) & 0.95 \\
    Learning rate & \text{manually scheduled} \\
    Clip range & 0.015 \\
    Entropy coefficient & 0.001 \\
    Value loss coefficient & 2.0 \\
    Batch size & 256 \\
    Epochs per update & 5 \\
    Steps per rollout & 2048 $\times$ \texttt{num\_parallel\_envs} \\
    Target KL & 0.3 \\
    number of parallel envs & 10 \\
    Weight decay & 0.01 \\
    total timesteps & 8e6 \\
    Advantage normalization & False \\
    \bottomrule
  \end{tabular}
\end{table}

\section*{ACKNOWLEDGMENTS}
The author thanks Yuval Tassa for answering their questions regarding anisotropic friction on MuJoCo's gitlab discussions, and for providing a useful fix.
\end{small}
\end{document}